\documentclass[letterpaper,10pt,conference]{ieeeconf}

\usepackage{amsmath}
\usepackage{hyperref}
\usepackage{tislbib}
\usepackage[nameinlink]{cleveref}
\usepackage[table]{xcolor}
\definecolor{tablehighlight}{HTML}{EFEFEC}
\usepackage{booktabs}
\usepackage{makecell}
\usepackage{graphicx}
\usepackage{multirow}
\usepackage[section]{placeins}

\IEEEoverridecommandlockouts
\hypersetup{colorlinks,linkcolor=uoftblue,urlcolor=uoftblue,citecolor=uoftblue}

\usepackage{amssymb}
\usepackage{uoftcolors}

\newcommand{\METRIC}{action margin}
\newcommand{\LANGSHARE}{language share}

\title{\LARGE\bf
When Listening Becomes Easier: Scrubbing Visual Cues\\ for Shortcut-Free VLAs
}

\author{Jasper Gerigk$^{1,2}$, Kenzo Aspuru-Takata$^{1}$, Chin-Hsuan Wu$^{1,2}$, Mohammad Mohammadi$^{1,2}$,\\ Shuhong
      Zheng$^{1,2}$, Igor Gilitschenski$^{1,2}$%
\thanks{*This work was supported by Vector Institute and Digital Research Alliance of Canada}%
\thanks{$^{1}$ University of Toronto, Toronto, ON M5S 3H5, Canada}
\thanks{$^{2}$ Vector Institute, Toronto, ON M5G 0C6, Canada}
}

\begin{document}

\maketitle
\thispagestyle{empty}
\pagestyle{empty}


\begin{abstract}
Shortcut learning is a prevalent issue in robot learning. 
The limited diversity of robot demonstration datasets can mislead policies into exploiting spurious correlations between tasks and irrelevant features, such as viewpoint or background.
Collecting sufficiently diverse robot demonstrations is costly and inefficient, motivating algorithmic alternatives.
We focus on vision-language-action (VLA) models and discover that different vision-language model backbones exhibit substantially different levels of susceptibility to visual shortcut learning. 
We find that model behavior correlates with our proposed representation-level metric, \textit{action margin}, which requires no policy rollouts.
Visual shortcuts consistently enter action representations in early layers, with models differing in the extent to which later layers correct them by incorporating language information.
To boost models' attention to language, we introduce task scrubbing, a new domain-adversarial training method that decreases models' likelihood of using visual shortcuts and improves VLAs' generalization. 
Experiments in both simulation and the real world across multiple VLAs and visual cues show that task scrubbing improves out-of-distribution robustness and often eliminates visual shortcut learning.
\end{abstract}

\section{Introduction}

Vision-language-action (VLA) models are general-purpose robot policies built on pretrained vision-language models~(VLMs)~\cite{alayrac2022flamingo,liu2023visual} for their visual and linguistic abilities~\cite{brohan2023rt,kim2024openvla}.
Although a large variety of different VLA models~\cite{black2024pi_0,shukor2025smolvla,zheng2026x, wang2026qwen} have been proposed, they share a common design:
a VLM encodes and processes visual and language information and produces representations which are then used to predict actions.
Actions are predicted either by the same backbone~\cite{kim2024openvla,belkhale2024minivla} or by a separate transformer, commonly called an action expert, that attends to the VLM representations.
This design has produced policies that transfer pretrained vision-language knowledge to diverse manipulation tasks and generalize beyond individual training environments~\cite{brohan2023rt,black2024pi_0,intelligence2025pi}.
Its appeal rests on an implicit premise: the policy should inherit sufficiently strong visual-language understanding from the VLM to condition its actions on the language instruction.

However, in practice, this premise is less reliable than benchmark performance suggests~\cite{jiang2026we,fei2025libero,xu2026unmasking}.
VLAs remain brittle to visual perturbations and initial-state changes~\cite{fei2025libero}, may fail to generalize across linguistic paraphrases~\cite{kim2026liberopara}, and can lose useful pretrained structure while adapting their representations to precise spatial and action demands~\cite{driess2025knowledge,grover2025enhancing,dalal2026generalizable}.
More generally, robot policies tend to rely on irrelevant features captured in their training distribution~\cite{xing2025shortcut}.
For example, proxies such as viewpoint, background, object color, distractors, or proprioceptive state may provide a strong signal for task identity in the limited diversity of the training data~\cite{swann2026sparse, grant2026not}.
This behavior is an instance of shortcut learning~\cite{geirhos2020shortcut,xing2025shortcut}.
Models affected by shortcut learning achieve high in-distribution (ID) success but follow inconsequential cues when determining the task under out-of-distribution~(OOD) conditions and thus fail.

\begin{figure}
    \centering
    \includegraphics[width=1\linewidth]{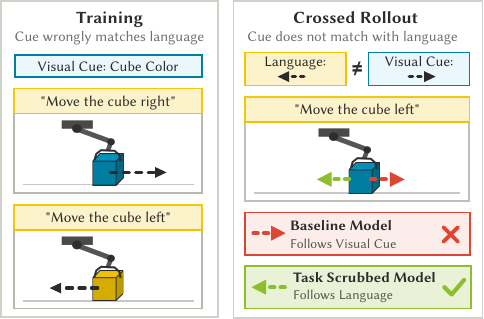}
    \caption{\textbf{Visual shortcut-learning setup.} During training (left), the model sees two different tasks with visual cues that match the language instruction. When the model is then evaluated (right) out-of-distribution where the visual cue does not match the language, the baseline model follows the visual cue rather than the language. Applying task scrubbing during training encourages the policy to follow the instruction.}
    \label{fig:overview}
    \vspace{-15pt}
\end{figure}

We focus on \emph{visual shortcuts}~(\Cref{fig:overview}): task identity-correlated visual features that the policy uses as substitutes for the language instruction. 
These shortcuts are especially common as small visual variations between tasks can easily enter training demonstrations through shifts in camera position, background changes, or task setup.
When this unwanted association is broken, a shortcut policy executes the visual cue-consistent behavior instead of the instructed behavior. 
Existing work largely describes this failure as a data-diversity problem~\cite{xing2025shortcut}, without studying why different VLA architectures trained on the same data can have different shortcut propensities.
It also remains unclear how shortcuts enter action computation or how the visual shortcut pathway can be suppressed without discarding useful visual information.

To understand how model architectures influences shortcut susceptibility, we evaluate multiple existing VLAs and our own variants to better understand the effect of VLM choice.
We find that shortcut susceptibility varies substantially across VLM backbones on the same dataset.
Further, we introduce the \METRIC~metric, which tracks whether a model is likely to shortcut by the alignment of the action-token hidden states with task- and cue-consistent reference states. 
Using this metric, we can directly track the development of shortcut actions during inference and find a mechanistic explanation. 
Attention knockout~\cite{geva2023dissecting} shows that task-correlated visual features shift action representations toward the shortcut in early layers with models differing in whether later computation restores alignment with the instruction.
Based on these findings, we introduce task scrubbing to encourage models to attend to language information.
Task scrubbing is an adversarial auxiliary loss that causes visual representations that predict task identity to shift during training, making them less reliable for action prediction.
The intervention increases the relative influence of language when the model learns to predict actions, without removing information from the visual input.

Our contributions are:
\begin{itemize}
    \item we show that shortcut susceptibility is strongly architecture dependent: VLA variants trained on the same data behave substantially differently, while VLM backbone size and general performance do not predict robustness;
    \item we introduce a systematic framework for analyzing shortcut behavior: \METRIC{} tracks behavioral shortcut reliance without rollouts, while attention knockout identifies how visual and language information shape action representations; and
    \item we introduce task scrubbing, a gradient-reversal method that makes task prediction from the visual pathway harder during training and causes policies to rely more strongly on language, thereby reducing shortcut learning.
\end{itemize}

\section{Background and Related Work}

\subsection{Vision-Language-Action Models}

VLAs map visual observations, language instructions, and, in many systems, proprioceptive state to robot actions~\cite{brohan2023rt,kim2024openvla}.
They vary in how they connect pretrained visual-language representations to control.
Some models predict tokenized actions autoregressively with the VLM backbone itself~\cite{kim2024openvla,belkhale2024minivla}, whereas others pass image and text representations to a separate action expert that generates continuous action chunks~\cite{black2024pi_0,shukor2025smolvla}.
Multi-pathway systems such as $\pi_{0.5}$ make this separation more explicit by assigning goal semantics and motor generation to different components~\cite{intelligence2025pi}.
Across these designs, the expected benefit is that the VLM supplies broad semantic and visual abstractions while the action component adapts them to motor execution.

Despite their successes, VLAs face general robustness and representation-learning challenges.
LIBERO-Plus~\cite{fei2025libero} documents brittleness under distractors, initial-state changes, and other visual variations, while LIBERO-Para shows that robustness to linguistic rephrasing cannot be assumed from VLM pretraining~\cite{kim2026liberopara}.
VLM features may omit fine spatial or action-relevant details, motivating adaptation during policy training~\cite{driess2025knowledge}.
However, broad fine-tuning can degrade pretrained visual-language structure and hurt generalization~\cite{grover2025enhancing, dalal2026generalizable}.
Freezing the pretrained VLM to prevent information loss does not generally help and can reduce task performance without improving language generalization~\cite{kim2026liberopara}.
VLAs therefore face a tension between adapting representations to predict actions accurately and preserving the general structure to enable generalization.
Various approaches have been proposed to improve representations, such as using stronger VLMs~\cite{wu2026vlanext}, utilizing more informative visual embeddings~\cite{miao2026jepa}, or adding representation-preserving objectives~\cite{driess2025knowledge, grover2025enhancing}. 
Larger systems such as $\pi_{0.7}$ combine broader data, model, and training changes to improve generalization~\cite{intelligence2026pi}.
These results suggest that architecture and training affect representation quality and robustness.

\subsection{Shortcut Learning in Robot Policy Training}

Shortcut learning occurs when a policy exploits a predictive correlation that is easier to learn than the intended task rule and consequently fails under distribution shift~\cite{geirhos2020shortcut}. 
In generalist robot policies, insufficient dataset diversity and fragmentation allow task identity to become entangled with particular scenes, embodiments, camera configurations, or
trajectories~\cite{xing2025shortcut}. 
More data reduces this problem only when it introduces the relevant variation.

Mechanistic studies find weak prompt control when visual appearance already determines the behavior and identify internal features associated with memorized trajectories~\cite{grant2026not,swann2026sparse}. 
These findings challenge the assumption that a successful VLA necessarily learns a compositional mapping from instruction and scene to action. 
The model may instead interpolate among visually indexed motor programs.

Existing mitigations attempt to break shortcut correlations through broader data collection~\cite{xing2025shortcut}.
Such interventions can be costly, benchmark-specific, and difficult to transfer.
Removing shortcut-prone observation channels can also improve generalization~\cite{zhao2025you}, but permanently discarding an input removes information that other tasks may require.
AC-VLA addresses visual shortcut reliance by selectively removing wrist-camera observations when the gripper is closed~\cite{peng2026ac}.
DeLock takes an alternative approach of recovering the pretrained knowledge after fine-tuning using a contrastive prompt and steering~\cite{huang2026breaking}.
We expand on prior analysis and investigate how the model's vision-language backbone impacts a policy's propensity to follow visual shortcuts.
Rather than permanently removing information or requiring additional data, we develop a training modification which causes VLAs to avoid relying on visual shortcuts.

\subsection{Adversarial Representation Learning}

Domain-adversarial representation learning~\cite{ganin2016domain} is a general approach to make representations from different data sources indistinguishable to prevent models from improperly relying on them when making decisions.
It relies on an auxiliary classifier to predict a selected attribute from a representation and reverses its gradient into the encoder~\cite{ganin2016domain}. 
The classifier learns to recover the attribute, while the encoder is trained to make that prediction difficult.
This mechanism is commonly used to promote invariance to domains or nuisance variables, particularly in computer vision~\cite{yang2024limits,feng2026rap}, but has not yet been applied in related robotics settings. 
While making two sources of data completely indistinguishable can be very difficult, we find that continually shifting task-predictive visual representations during training can be sufficient to make the policy rely on the instruction rather than the visual shortcut.  

\section{Problem Formulation and Analysis}

\subsection{Formal Definition of Shortcut Learning}

We first establish a formal definition of visual shortcut learning.
Let $T$ denote the task specified by language, $U$ denote the task-relevant visual information needed to execute the task, $Z$ denote a task-irrelevant visual cue, and $A$ denote the desired action. 
The complete visual observation is $X=(U,Z)$. 
In the training distribution, task and cue identity are statistically dependent such that $I_{\mathrm{train}}(T;Z)>0.$
In our case, shortcut learning refers to cases where the learned policy mistakenly relies on $Z$ rather than $T$.

The intended policy uses the instruction and task-relevant visual information: $\pi_{\mathrm{intended}}(A\mid U,T)$.
A shortcut policy instead allows the irrelevant cue to substitute for the instruction: $\pi_{\mathrm{shortcut}}(A\mid U,Z)$.
Its behavior depends strongly on $Z$ and is comparatively invariant to $T$, even though $Z$ is not causally relevant to the desired action.

Operationally, training contains correlated pairs such as $(T_A,Z_A)$ and $(T_B,Z_B)$. 
In-distribution~(ID) evaluation retains these observed combinations, whereas out-of-distribution~(OOD) evaluation crosses them:
\begin{equation} \label{eq:ood_states}
    (T_A,Z_B), \qquad (T_B,Z_A).
\end{equation}
A robust policy follows the language-specified task under these crossings; a shortcut policy follows the visual cue.

We report ID success on the task--cue combinations present during training and OOD success on crossed combinations with success corresponding to the standard task completion criterion of the environment. 
We additionally report shortcut rate~(SC), defined as the percentage of all OOD episodes in which the policy attempts the cue-consistent~($Z$) task rather than the instructed task~($T$). 
An OOD episode cannot both succeed and shortcut.

\begin{figure}[t]
    \centering
    \vspace{5pt}
    \includegraphics[width=\linewidth]
        {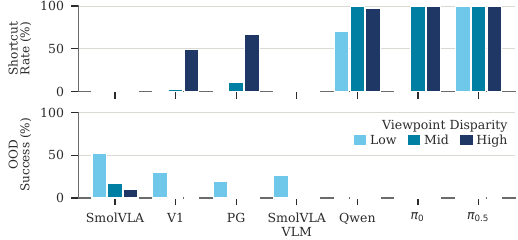}
    \vspace{-15pt}
    \caption{OOD success and SC across VLA variants and low, medium, and high task-correlated viewpoint disparity on LIBERO Spatial.}
    \label{fig:vp_disp_results}
\end{figure}

\subsection{LIBERO Spatial Viewpoint-Disparity Setup}
\label{sec:lib_spatial_viewpoint_results}

We adopt the controlled viewpoint-disparity setup of prior shortcut-learning work~\cite{xing2025shortcut}, where we have a dataset of two LIBERO Spatial~\cite{liu2023libero} tasks with task-specific viewpoints.
The Low, Mid, and High conditions use progressively larger viewpoint differences between the two tasks.
Across a broader set of VLAs, we reproduce the reported shortcut behavior~(\Cref{fig:vp_disp_results}).
A rollout is considered to have used a shortcut if the gripper is closer to the incorrect bowl location when closing the first time.
We compare finetuned $\pi_0$~\cite{black2024pi_0}, $\pi_{0.5}$~\cite{intelligence2025pi}, SmolVLA~\cite{shukor2025smolvla}, and SmolVLA-VLM (SmolVLA with randomly initialized action-expert weights).
We further introduce and evaluate novel SmolVLA variants, which use the same architecture but use a different VLM, to better understand the impact of the chosen backbone. 
SmolVLA-V1 replaces SmolVLA's SmolVLM2 backbone with SmolVLM1, SmolVLA-PG uses the PaliGemma backbone~(used by $\pi_0$), and SmolVLA-Qwen uses
Qwen2.5-VL-3B-Instruct~\cite{bai2025qwen25vltechnicalreport}. 

Using the larger variety of models, we can better identify patterns between various VLM backbones.
As shown in \Cref{fig:vp_disp_results}, SmolVLA and SmolVLA-VLM do not show any shortcut behavior, despite being significantly smaller models than $\pi_0$ or $\pi_{0.5}$.
The action pretraining further allows SmolVLA to achieve nonzero OOD success at all three disparity levels. 
Replacing its VLM backbone generally worsens performance, with the PaliGemma variant showing the strongest shortcut susceptibility.
Further, while Qwen2.5-VL~\cite{bai2025qwen25vltechnicalreport} has the highest average benchmark performance of the evaluated VLMs, the resulting VLA has a low success rate and high SC.
This indicates that general VLM performance does not predict the resulting VLA's propensity for shortcut learning.
The varying shortcut rates show that shortcut learning is not determined by the dataset alone and that both the VLM backbone and action pretraining matter.

\begin{figure}[t]
    \centering
    \includegraphics[width=0.95\linewidth]
        {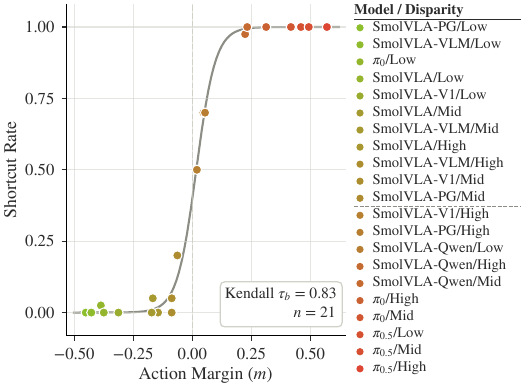}
    \caption{Action margin shows a strong correlation with shortcut rate for the 21 model--disparity combinations. The representation-level metric strongly tracks shortcut reliance (Kendall's $\tau_b=0.83$, $p=1.35\times10^{-7}$).}
    \label{fig:rel_marg_vs_sc}
\end{figure}

\begin{figure*}[t]
    \centering
    \vspace{5pt}
    \includegraphics[width=\linewidth]{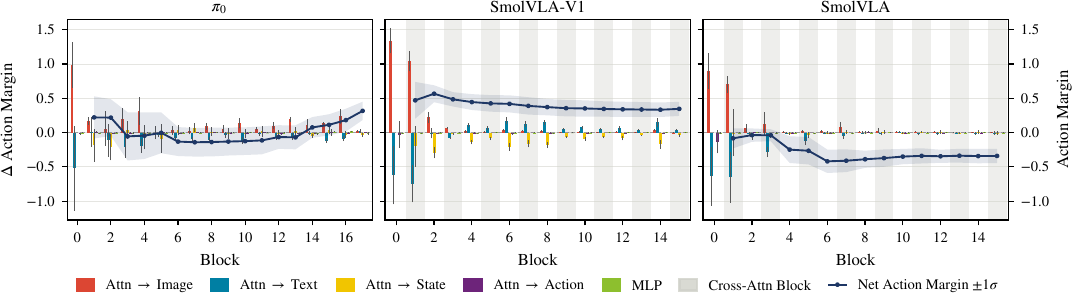}
    \caption{Layer-wise attention-knockout effects for $\pi_0$, SmolVLA-V1, and SmolVLA. 
    A positive value indicates that the removed attention promotes cue alignment, while negative values indicate it promotes task alignment.
    Removing visual attention generally shifts representations away from the cue, while language and action-token interactions can correct cue alignment in later layers.}
    \vspace{-15pt}
    \label{fig:a13}
\end{figure*}

\subsection{Representation-Level Shortcut Magnitude}

To estimate a model's shortcut susceptibility without policy rollouts, we introduce \METRIC{} as a proxy, which measures whether action representations of crossed OOD states (\Cref{eq:ood_states}) align more closely with the instructed task or the visual cue.
For each model, we use the average across the first observations of all training trajectories.
We average the per-dimension normalized final-layer action-token hidden states $h(o,\ell)$ for observation $o$ under instruction $\ell$.
We define the centroid representations $c_{T,Z} = \mathbb{E}_{o \in O(Z)}[h(o, \ell_T)]$, where $T$ is the task and $O(Z)$ is the set of visual observations which include visual cue $Z$ for the two ID pairings ($c_{T_A,Z_A}$ and $c_{T_B,Z_B}$).
We can then create an OOD sample such as $x=(T_A,Z_B)$.
For this pairing, $c_{\mathrm{task}}=c_{T_A,Z_A}$ shares the instruction-specified task, whereas $c_{\mathrm{cue}}=c_{T_B,Z_B}$ shares the visual cue.
Action margin $m(x)$ is defined as the relative distances to these two representations and can be used to determine if the predicted action follows the task or the visual cue. 
\begin{equation}
    m(x) = \frac{\lVert h(x) - c_{\mathrm{task}} \rVert_2 - \lVert h(x) - c_{\mathrm{cue}} \rVert_2}
                {\lVert c_{\mathrm{task}} - c_{\mathrm{cue}} \rVert_2}
    \label{eq:action_margin}
\end{equation}
Positive values indicate that the representation is closer to the cue-matched centroid, while negative values indicate closer alignment to the instructed task.

We obtain a model-level value by averaging $m(x)$ across the first observations of both crossed configurations.
Restricting the analysis to the first step and fixing the flow-matching noise puts all trajectories at the same stage of motion and avoids confounding the representation with divergent rollout states.
Across 21 model--viewpoint disparity combinations, \METRIC{} strongly correlates with SC (\Cref{fig:rel_marg_vs_sc}), demonstrating that it is a good indicator of shortcut reliance across architectural variation.

\subsection{Mechanistic Analysis}
\label{sec:mech_anal_lang_share}

Using \METRIC{} as a proxy for shortcut behavior, we apply attention knockouts~\cite{geva2023dissecting} to determine which input streams causally move the action representation toward or away from the visual cue.
At a selected layer, attention knockout masks all attention from action-token queries to one source--image, language, state, or other action tokens--before softmax renormalization.
We compute the impact as $\Delta m = m_{\mathrm{full}} - m_{\mathrm{knockout}}$ and average the behavior across ten initial trajectory frames.
The results for all layers are plotted in \Cref{fig:a13} for $\pi_0$, SmolVLA-V1, and SmolVLA. 
A positive $\Delta m$ indicates that removing a stream reduces cue alignment (the stream promotes the shortcut), whereas a negative $\Delta m$ indicates that the stream actively corrects the shortcut.

\Cref{fig:a13} shows that visual information injects the shortcut in early layers, consistently moving representations toward the cue.
Later correction is model-dependent: language and action-token interactions can steer representations back toward the instructed task, but this correction is weak or even reversed in shortcut-prone models.
Among the tested models, SmolVLA most effectively reverses early visual alignment via language features.
These results motivate increasing the relative influence of language during training.

As the language correction is desirable, we track this influence using \textit{\LANGSHARE}, which measures the relative extent to which language and visual cues steer the action representation along the task axis.
As before, let $h(o,\ell)$ denote the averaged final layer action-token representation, and let $(T_A,Z_A)$ and $(T_B,Z_B)$ denote the two ID combinations, with instructions $\ell_{T_A},\ell_{T_B}$. 
We define the task axis as the normalized difference of the two ID centroids,
\begin{equation}
    u=\frac{c_{T_A,Z_A}-c_{T_B,Z_B}}{\lVert c_{T_A,Z_A}-c_{T_B,Z_B}\rVert_2},
\end{equation}
and project two counterfactual displacements onto it. 
Re-running every observation under both instructions, with the image held fixed, gives the language steering
\begin{equation}
    S_L=\Big\langle \mathbb{E}_{o\in (O(Z_A)\bigcup O(Z_B))}\big[h(o,\ell_{T_A})-h(o,\ell_{T_B})\big],\,u\Big\rangle.
\end{equation} 
Holding the instruction fixed and contrasting the two visual cues gives the cue steering
\begin{equation}
    S_V=\Big\langle \mathbb{E}_{\ell\in\{\ell_{T_A},\ell_{T_B}\}}\big[\underset{O(Z_A)}{\mathbb{E}}[h(o,\ell)] - \underset{O(Z_B)}{\mathbb{E}}[h(o,\ell)]\big],\,u\Big\rangle .
\end{equation}
The diagnostic is their normalized balance,
\begin{equation}
    \text{\LANGSHARE}=\frac{\lvert S_L\rvert}{\lvert S_L\rvert+\lvert S_V\rvert}\in[0,1],
    \label{eq:language_share}
\end{equation}
which rises toward $1$ as language assumes control of the task axis and sits near $0$ for a pure-shortcut policy.
As in \Cref{eq:action_margin}, the metric is computed on the first frame of each trajectory under fixed flow-matching noise, and allows us to track the model's behavior during training.

\begin{figure}[t]
    \centering
    \vspace{5pt}
    \includegraphics[width=1\linewidth]{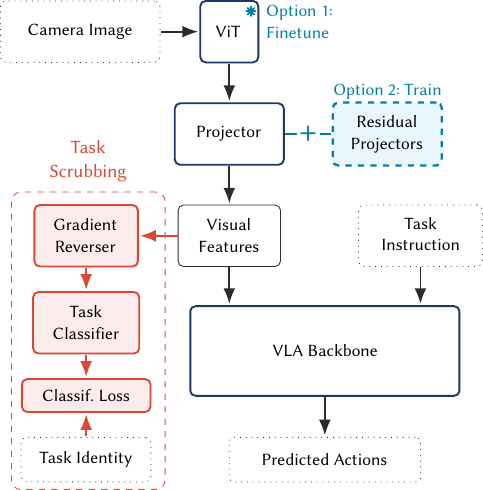}
    \caption{
    Task scrubbing makes task information less available in the visual features during training, so that the VLA backbone learns to pay more attention to the language information.
    Red denotes the auxiliary task-scrubbing head used during training, which makes the visual pathways less informative about task identity.
    Either option shown in blue allows the visual pathway to be trainable with little additional cost. 
    }
    \label{fig:method}
    \vspace{-15pt}
\end{figure}

\section{Task Scrubbing}

Task scrubbing is a training-time intervention that makes visually encoded task identity less stable and thus less convenient for the model to exploit during policy training.
As summarized in \Cref{fig:method}, it attaches an adversarial task classifier to the visual pathway: the classifier learns to recover task identity, while the visual pathway learns to frustrate that classifier.
Because the resulting visual representations continually change, they provide the policy with a less stable signal for task identity.
Language therefore becomes the more reliable task-identifying signal for the action objective.
To summarize, the goal is not to remove all task-decodable visual information, which is infeasible when task views differ strongly, but to prevent the task-irrelevant cues from becoming the easiest basis for action selection.

The classifier operates on the projected visual tokens before they enter the language-model or action backbone.
Let $v_1,\ldots,v_N$ denote these tokens across all camera views.
We first mean-pool them,
$
    \bar{v}=\frac{1}{N}\sum_{i=1}^{N}v_i,
$
and then apply LayerNorm and a two-layer MLP with a GELU~\cite{hendrycks2016gelu} nonlinearity to predict the task label associated with the images.
A gradient-reversal layer connects $\bar{v}$ to this classifier.

The total training objective is
\begin{equation}
    \mathcal{L} = \mathcal{L}_{\mathrm{action}} + w\mathcal{L}_{\mathrm{task}}, \label{eq:task_scrub_loss}
\end{equation}
where $\mathcal{L}_{\mathrm{task}}$ is cross-entropy on the task identity and $w$ sets the relative scale of the auxiliary adversarial signal.
This is a model- and setting-specific hyperparameter.
In the forward pass, the classifier predicts task identity normally.
During backpropagation, it minimizes $\mathcal{L}_{\mathrm{task}}$, while gradient reversal causes the visual pathway to maximize the same objective.
The classifier and policy are trained jointly from the beginning, and gradient reversal is applied throughout training.

For the auxiliary loss to affect the policy, the reversed gradient must reach trainable parameters in the visual pathway.
When the ViT is fine-tuned, task scrubbing can update it directly.
SmolVLA normally freezes both its VLM backbone and visual encoder, so we instead add a trainable residual adapter to the visual projection for such models.
If the original projection is $p(x)$, the adapted output is
\begin{equation}
    \tilde{p}(x) = p(x)+f_2\!\left(\operatorname{GELU}(f_1(x))\right),
    \label{eq:residual_adapter}
\end{equation}
where $f_1$ and $f_2$ are trainable linear layers with $f_2$ zero-initialized.
Thus task scrubbing adds negligible computational overhead to training.

\begin{figure}
    \centering
    \vspace{5pt}
    \includegraphics[width=1\linewidth]{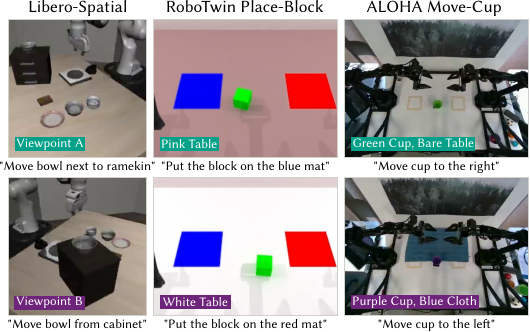}
    \caption{Evaluation setup: The three different environments and task pairs used for evaluation.}
    \label{fig:eval_tasks}
\end{figure}

\section{Results}

We evaluate task scrubbing on three separate experimental setups~(\Cref{fig:eval_tasks}), involving three types of visual shortcut.

\subsection{Task Scrubbing Improves LIBERO Robustness}

Returning to the setup from~\Cref{sec:lib_spatial_viewpoint_results}, we evaluate task scrubbing on the high-disparity LIBERO~\cite{liu2023libero} setting, where the two viewpoints differ by the largest rotation and the visual shortcut is the strongest.
We compare the baseline policy with task scrubbing and report mean and sample standard deviations across three seeds.
Each seed uses 40 rollouts for each of the four task--viewpoint configurations.
\Cref{tab:libero_results} reports ID and OOD success and SC.
The auxiliary-loss weight used for each model is reported in the Appendix.

\begin{table}[t]
\centering
\vspace{5pt}
\caption{LIBERO Spatial results under high viewpoint disparity.}
\label{tab:libero_results}
\begin{tabular}{llrrr}
\toprule
\thead{Model} & \thead{Condition} &
\thead{OOD\\Succ. (\%)} & \thead{SC (\%)} &
\thead{ID\\Succ. (\%)} \\
\midrule
& Baseline & $10 \pm 1$ & $0 \pm 0$ & $62 \pm 3$ \\
\multirow{-2}{*}{SmolVLA~\cite{shukor2025smolvla}} & Scrubbed & $73 \pm 3$ & $0 \pm 0$ & $74 \pm 4$ \\
\rowcolor{tablehighlight}
& Baseline & $4 \pm 6$ & $45 \pm 26$ & $37 \pm 4$ \\
\rowcolor{tablehighlight}
\multirow{-2}{*}{SmolVLA-V1} & Scrubbed & $64 \pm 9$ & $0 \pm 0$ & $66 \pm 3$ \\
 & Baseline & $4 \pm 5$ & $17 \pm 28$ & $34 \pm 8$ \\
\multirow{-2}{*}{SmolVLA-VLM} & Scrubbed & $64 \pm 10$ & $0 \pm 0$ & $66 \pm 10$ \\
\rowcolor{tablehighlight}
& Baseline & $0 \pm 1$ & $100 \pm 1$ & $50 \pm 5$ \\
\rowcolor{tablehighlight}
\multirow{-2}{*}{MiniVLA~\cite{belkhale2024minivla}} & Scrubbed & $26 \pm 13$ & $0 \pm 0$ & $44 \pm 5$ \\
& Baseline & $0 \pm 0$   & $99 \pm 1$  & $70 \pm 9$ \\
\multirow{-2}{*}{$\pi_0$~\cite{black2024pi_0}} & Scrubbed & $48 \pm 17$ & $0 \pm 0$   & $69 \pm 7$ \\
\rowcolor{tablehighlight}
& Baseline & $0 \pm 0$   & $100 \pm 0$ & $66 \pm 4$ \\
\rowcolor{tablehighlight} \multirow{-2}{*}{$\pi_{0.5}$~\cite{intelligence2025pi}} & Scrubbed & $28 \pm 25$ & $28 \pm 46$ & $45 \pm 4$ \\
\bottomrule
\end{tabular}
\end{table}

Task scrubbing reduces SC and increases OOD success for $\pi_0$, the SmolVLA family, and MiniVLA~\cite{belkhale2024minivla}. 
For $\pi_0$, OOD success rises from 0\% to 48\% while SC falls from 99\% to 0\%, with no meaningful reduction in ID success.
The main exception is $\pi_{0.5}$.
Task scrubbing reduces mean SC and increases mean OOD success, but succeeds for only two of three seeds.
The remaining seed retains an SC of approximately 90\%.
Mean ID success also falls from 66\% to 45\%.
The SmolVLA variants show the strongest overall response: observed shortcut behavior is eliminated, and ID performance increases.

\subsection{Task Scrubbing Alters the Internal Information Pathway}

During training, the accuracies of both the adversarial classifier and an additional gradient-isolated probe with the same architecture fall to chance or below. 
However, linear probes trained after policy training can still consistently decode task identity.
Thus task scrubbing does not fully remove task information from the image stream.
Instead the adversarial objective continually changes the task-relevant visual representation, providing the action learner with less stable visual signals for task identity.
Language remains a direct and stable source of task identity. 
Consequently, task scrubbing changes which pathway is easiest to use rather than permanently deleting information from the image.

\begin{figure}[t]
    \centering
    \vspace{5pt}
    \includegraphics[width=\linewidth]{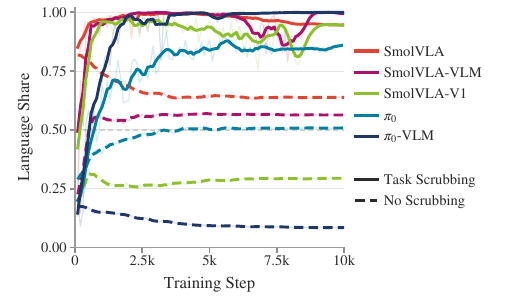}
    \caption{Language share during training for SmolVLA, SmolVLA-VLM, SmolVLA-V1, $\pi_0$, and $\pi_0$ without action pretraining. 
    Task scrubbing increases the relative influence of language on the action representations. 
    Action pretraining raises initial \LANGSHARE{} in both the SmolVLA and $\pi_0$ comparisons, but pretrained models respond more slowly to scrubbing.}
    \label{fig:lang_share}
\end{figure}

Training runs shown in \Cref{fig:lang_share} show this routing effect. 
Without task scrubbing, \LANGSHARE{} does not increase sufficiently or even decreases during training.
With scrubbing, it rises sharply, indicating that action representations become more sensitive to language when determining the task information. 
Initial values differ across architectures: SmolVLA begins with a comparatively high \LANGSHARE{}, and action pretraining increases the initial value in both the SmolVLA and $\pi_0$ comparisons, indicating that pretraining on a diverse set of tasks helps the models learn to use language. 

\begin{table}[t]
\centering
\vspace{5pt}
\caption{Place-Block results with different backgrounds.}
\label{tab:robotwin_results}
\renewcommand{\arraystretch}{1.15}
\begin{tabular}{llrrr}
\toprule
\thead{Model} & \thead{Condition} &
\thead{OOD\\Succ. (\%)} & \thead{SC (\%)} &
\thead{ID\\Succ. (\%)} \\
\midrule
\multirow{2}{*}{SmolVLA~\cite{shukor2025smolvla}}
    & Baseline & $5 \pm 2$ & $0 \pm 0$ & $98 \pm 3$ \\
    & Scrubbed & $86 \pm 3$ & $0 \pm 0$ & $95 \pm 2$ \\
\rowcolor{tablehighlight}
    & Baseline & $0 \pm 0$ & $100 \pm 0$ & $90 \pm 11$ \\
\rowcolor{tablehighlight} \multirow{-2}{*}{SmolVLA-V1}
    & Scrubbed & $83 \pm 6$ & $0 \pm 0$ & $89 \pm 5$ \\
\multirow{2}{*}{SmolVLA-VLM}
    & Baseline & $0 \pm 0$ & $100 \pm 0$ & $95 \pm 2$ \\
    & Scrubbed & $45 \pm 15$ & $0 \pm 0$ & $50 \pm 11$ \\
\rowcolor{tablehighlight}
    & Baseline & $0 \pm 0$ & $100 \pm 0$ & $94 \pm 1$ \\
\rowcolor{tablehighlight} \multirow{-2}{*}{$\pi_0$~\cite{black2024pi_0}}
    & Scrubbed & $62 \pm 12$ & $0 \pm 0$ & $85 \pm 0$ \\
    & Baseline & $0 \pm 0$ & $100 \pm 0$ & $97 \pm 1$ \\
\multirow{-2}{*}{$\pi_{0.5}$~\cite{intelligence2025pi}}
    & Scrubbed & $39 \pm 8$ & $39 \pm 9$ & $57 \pm 18$ \\
\bottomrule
\end{tabular}
\end{table}

\subsection{RoboTwin}

We next evaluate whether the LIBERO results transfer to a different simulator and cue type.
We construct a Place-Block task in the RoboTwin 2.0~\cite{chen2025robotwin} simulator, which uses an ALOHA~\cite{Zhao-RSS-23} robot.
The task requires grasping a cube at the center and placing it either on a blue mat to the left or on a red mat to the right using the corresponding arm according to the language instruction.
During training, a brown background is paired with the left-placement task and a white background with the right-placement task. 
We train on a dataset of 400 episodes, 200 of each task, generated using RoboTwin's data generator.

Task success is measured by whether the cube is fully inside the instructed region.
SC is measured by whether the gripper which closes closest to the cube is associated with the cue-consistent task. 
We evaluate each of the four cube-background color combinations 20 times and report standard deviations across three seeds.
\Cref{tab:robotwin_results} shows the same overall pattern as LIBERO~(see Appendix for the loss weights used).
We do not evaluate MiniVLA as its ID performance is too low to support a meaningful comparison.
SmolVLA-VLM has a similar pattern to $\pi_{0.5}$ on LIBERO Spatial, where one seed performs significantly worse than the other two.
Models more often see a drop in ID performance, possibly as this cue is stronger than the viewpoint change.
But task scrubbing still decreases SC and boosts OOD success.

\begin{table}[t]
\centering
\vspace{5pt}
\caption{Real-world ALOHA performance.}
\label{tab:real_world_results}
\renewcommand{\arraystretch}{1.15}
\resizebox{\linewidth}{!}{%
\begin{tabular}{lllrrr}
\toprule
\rowcolor{white}\thead{Cue} & \thead{Model} & \thead{Scrub.?} & \thead{OOD \\ Succ. \\ (\%)} & \thead{SC \\ (\%)} & \thead{ID \\ Succ. \\ (\%)} \\
\midrule
 & \multirow{2}{*}{SmolVLA~\cite{shukor2025smolvla}} & No & 77.5 & 0 & 95.0 \\
 & & Yes & 95.0 & 2.5 & 95.0 \\
\rowcolor{tablehighlight}\cellcolor{white!10}
 &  & No & 57.5 & 17.5 & 95.0 \\
\cellcolor{tablehighlight}\cellcolor{white!10}\multirow{-4}{*}{\makecell{Cup color}} & \multirow{-2}{*}{\cellcolor{tablehighlight}SmolVLA-V1}& \cellcolor{tablehighlight}Yes & \cellcolor{tablehighlight}87.5 & \cellcolor{tablehighlight}0 & \cellcolor{tablehighlight}90.0 \\ 
\midrule
 & \multirow{2}{*}{SmolVLA~\cite{shukor2025smolvla}} & No & 47.5 & 27.5 & 100.0 \\
 & & Yes & 92.5 & 0 & 97.5 \\
\rowcolor{tablehighlight}\cellcolor{white!10}
 &  & No & 0 & 100.0 & 100.0 \\
\cellcolor{white!10}
 \multirow{-4}{*}{\makecell{+Back-\\ground}} & \cellcolor{tablehighlight}\multirow{-2}{*}{SmolVLA-V1}& \cellcolor{tablehighlight}Yes & \cellcolor{tablehighlight}100.0 & \cellcolor{tablehighlight}0 & \cellcolor{tablehighlight}97.5 \\ 
\bottomrule
\end{tabular}
}
\end{table}

\subsection{Real-Robot Experiments}

Finally, we evaluate task scrubbing on a physical ALOHA~\cite{Zhao-RSS-23} platform controlled through the Trossen LeRobot setup.
The robot must place a cup on either the left or right target according to a language instruction.
We test the models across two different visual shortcut conditions.
The first condition uses the cup color as a local cue. 
During training, a green cup is paired with the right target and a purple cup with the left target.
The second condition introduces an additional stronger global cue: the purple-left demonstrations also contain a blue towel in the background, while the green-right demonstrations use the normal table background. 
This condition therefore correlates task identity with both cup color and background. 
SC is recorded when the gripper associated with the cue-consistent task closes in an attempt to grasp the cup.

We collect 100 demonstrations per task and evaluate one trained seed per condition with twenty trials per configuration.
\Cref{tab:real_world_results} shows that task scrubbing improves both models under both cue conditions. 
We use the same weights as in the LIBERO Spatial results~(see Appendix).
With cup color alone, little shortcut learning occurs but models' OOD success still benefits from task scrubbing.
In the combined background-and-color condition, task scrubbing eliminates all observed shortcut episodes and yields OOD success rates above 90\% for both models.
These results show that the intervention transfers to physical control and is effective against both local and global visual cues.

\subsection{Ablation: Direct Attention Regularization}

Because task scrubbing increases \LANGSHARE{}, an alternative is to manipulate attention directly to produce a similar effect. 
To the action loss, we add an attention regularizer equal to the mean of the action-to-image attention over all heads and layers containing cross-modal attention:
\begin{equation}
    \mathcal{L}_{\mathrm{attn}} = \lambda\,
    \operatorname{mean} \left(A_{\mathrm{action}\rightarrow\mathrm{image}}\right), \qquad \lambda=1.
    \label{eq:attention_regularizer}
\end{equation}
Minimizing this term penalizes action-to-image attention relative to the available attention budget and drives \LANGSHARE{} to above 90\% in all evaluated models.

We compare attention regularization with task scrubbing on high-disparity LIBERO. 
\Cref{tab:tau-vs-dann} reports means and standard deviations.
Attention regularization works for the SmolVLA variants, but trails task scrubbing in success rates. 
It fails for the $\pi$ family: $\pi_0$ retains 68\% SC and $\pi_{0.5}$ retains 79\% SC. 
Thus, forcing a high language attention ratio does not guarantee that language controls the action as attention amounts are an incomplete measure of information flow.

\begin{table}
\centering
\vspace{5pt}
\caption{Attention regularization versus task scrubbing.}
\label{tab:tau-vs-dann}
\resizebox{\linewidth}{!}{%
\begin{tabular}{llccc}
\toprule
Model & Method &
\thead{OOD Succ.\\(\%)} & \thead{SC\\(\%)} &
\thead{ID Succ.\\(\%)} \\
\midrule
\multirow{2}{*}{SmolVLA~\cite{shukor2025smolvla}}
 & Scrubbed & $73 \pm 3$ & $0 \pm 0$ & $74 \pm 4$ \\
 & Attention Reg. & $67 \pm 4$ & $0 \pm 0$ & $67 \pm 3$ \\
\rowcolor{tablehighlight}
 & Scrubbed & $64 \pm 9$ & $0 \pm 0$ & $66 \pm 3$ \\
\rowcolor{tablehighlight}
\multirow{-2}{*}{SmolVLA-V1} & Attention Reg. & $56 \pm 5$ & $0 \pm 0$ & $59 \pm 4$ \\
\multirow{2}{*}{$\pi_0$~\cite{black2024pi_0}}
 & Scrubbed & $48 \pm 17$ & $0 \pm 0$ & $69 \pm 7$ \\
 & Attention Reg. & $3 \pm 2$ & $68 \pm 31$ & $57 \pm 8$ \\
\rowcolor{tablehighlight}
 & Scrubbed & $28 \pm 25$ & $28 \pm 46$ & $45 \pm 4$ \\
\rowcolor{tablehighlight}
\multirow{-2}{*}{$\pi_{0.5}$~\cite{intelligence2025pi}} & Attention Reg. & $0 \pm 0$ & $79 \pm 10$ & $50 \pm 10$ \\
\bottomrule
\end{tabular}
}
\end{table}

\section{Discussion}

Our results show that VLM choice changes how readily task-correlated visual information controls a VLA's action component. 
Models trained on the same demonstrations can differ substantially in shortcut behavior, and replacing SmolVLA's backbone is sufficient to change its robustness.
Shortcut learning should therefore not be treated solely as a dataset property. 
Architecture, pretraining, and the ability of downstream layers to use language to correct visually induced cue alignment are also important.

The mechanistic analysis further shows that robustness does not require a model to stop encoding the cue.
Visual task identity remains decodable after task scrubbing. 
What matters is whether that information dominates action computation and whether later layers use language information to restore task alignment.
Task scrubbing changes this competition by making visually encoded task identity unstable during training. 
The resulting increase in \LANGSHARE{} is evidence of rerouting toward language, rather than complete
removal of visual information.

The proposed diagnostics may also support model selection and training monitoring.
Under controlled crossings, the \METRIC{} can identify backbones whose action representations are likely to follow the cue. 
Language share can track whether task scrubbing shifts representational influence toward language during training and may provide a signal for adjusting $w$. 

\section{Conclusion and Limitations}

Using \METRIC{} together with attention knockout, we show that visual cues shift early-layer VLA action representations toward cue-consistent behavior.
We observe differences in how later computation reflects language control across VLM backbones, which can be tracked during training using \LANGSHARE{}.
Task scrubbing makes visually encoded task identity unstable during training, causing the policy to prefer the language instruction as its task-identifying signal. 
Across LIBERO, RoboTwin, and real-robot experiments, it reduces shortcut behavior and improves OOD success. 
Overall, our findings establish that directly manipulating information routing is an effective way to prevent visual shortcut learning in VLAs.

Nevertheless, our approach still has limitations.
Task scrubbing requires task labels and a visual pathway with trainable parameters.
Frozen architectures therefore require an adapter.
It also requires architecture-specific calibration of the auxiliary adversarial signal; developing a reliable ID-only calibration criterion remains open.
Our mechanistic analysis focuses on VLM-plus-action-expert architectures, and MiniVLA is consequently included only in the behavioral intervention results, as supporting its autoregressive nature would require architecture-specific methodology.
Our evaluations are limited to settings containing two paired tasks and do not investigate how task scrubbing behaves in multi-task settings.
Finally, some seeds retain shortcut behavior, and task scrubbing can reduce ID performance. 

\section*{Appendix}

The task scrub scale $w$ was selected for each architecture and task.
Using a single seed, we select a scale that reduces SC and maximizes OOD success without reducing ID success when possible.
\Cref{tab:auxiliary_loss_scales} shows the values used.

\begin{table}[ht]
\centering
\caption{Selected auxiliary-loss scales for task-scrubbing.}
\label{tab:auxiliary_loss_scales}
\begin{tabular}{lrrr}
\toprule
\thead{Model} & \thead{LIBERO} & \thead{RoboTwin} & \thead{Real robot} \\
\midrule
SmolVLA~\cite{shukor2025smolvla} & 2 & 0.75 & 2 \\
SmolVLA-V1 & 1 & 0.175 & 1 \\
SmolVLA-VLM & 2 & 0.25 & -- \\
MiniVLA~\cite{belkhale2024minivla} & 12 & -- & -- \\
$\pi_0$~\cite{black2024pi_0} & 0.5 & 0.35 & -- \\
$\pi_{0.5}$~\cite{intelligence2025pi} & 0.8 & 0.55 & -- \\
\bottomrule
\end{tabular}
\vspace{-10pt}
\end{table}

\section*{Acknowledgments}

AI-assisted coding tools, including agents and coding harnesses, were used during development of the code.


\printbibliography

\end{document}